\documentclass[a4paper,11pt]{article}

\usepackage{geometry}
\usepackage{titling}
\usepackage[utf8]{inputenc}
\DeclareUnicodeCharacter{2212}{-}
\usepackage{amsmath}
\usepackage{microtype}
\usepackage{siunitx}
\usepackage{lineno}
\usepackage{authblk}
\usepackage{lscape}
\usepackage{graphicx}
\usepackage{subcaption}
\usepackage[font={footnotesize},labelfont=bf]{caption}
\usepackage[super,numbers,sort&compress]{natbib}
\usepackage{booktabs}
\usepackage{blindtext}
\usepackage{booktabs,multirow,array}
\usepackage{tikz}
\usepackage{verbatim}
\usepackage{chemfig}
\usepackage{gensymb}
\usepackage{textcomp}
\usepackage{setspace}
\usepackage{multicol}
\usepackage{filecontents,pgfplots,pgfplotstable}
\usepackage{afterpage}
\usepackage{xtab,afterpage}
\usepackage{amssymb}
\usepackage{rotating}
\usepackage{scrextend}
\usepackage{tcolorbox}
\usepackage{xr}
\usepackage[hidelinks]{hyperref}
\usepackage{cleveref}
\usepackage{algorithm}
\usepackage[noend]{algpseudocode}
\usepackage{enumitem}
\setlist[itemize]{leftmargin=*}
\usepackage{microtype}
\usepackage{float}

\algnewcommand\algorithmicinput{\textbf{Preprocessing:}}
\algnewcommand\Preprocessing{\item[\algorithmicinput]}

\usepackage{arxiv}

\usepackage{txfonts}

\begin{document}


\title{Two approaches to inpainting microstructure with deep convolutional generative adversarial networks}

\author[1]{{Isaac Squires}
\thanks{i.squires20@imperial.ac.uk}\ \ }

\author[1]{{Samuel J. Cooper \thanks{samuel.cooper@imperial.ac.uk}}\ \ }

\author[1]{{Amir Dahari}}

\author[1]{{Steve Kench}}

\affil[1]{{\textit{\footnotesize Dyson School of Design Engineering, Imperial College London, London SW7 2DB}}}

\lhead{\scshape Squires \textit{et al.}}
\chead{\scshape Microstructure Inpainting}
\rhead{\scshape Preprint}

\maketitle

\begin{abstract}
    {Imaging is critical to the characterisation of materials. However, even with careful sample preparation and microscope calibration, imaging techniques are often prone to defects and unwanted artefacts. This is particularly problematic for applications where the micrograph is to be used for simulation or feature analysis, as defects are likely to lead to inaccurate results. Microstructural inpainting is a method to alleviate this problem by replacing occluded regions with synthetic microstructure with matching boundaries. In this paper we introduce two methods that use generative adversarial networks to generate contiguous inpainted regions of arbitrary shape and size by learning the microstructural distribution from the unoccluded data. We find that one benefits from high speed and simplicity, whilst the other gives smoother boundaries at the inpainting border. We also outline the development of a graphical user interface that allows users to utilise these machine learning methods in a `no-code' environment. }
\end{abstract}

\section{Introduction}
\label{sec1}

Characterising materials with imaging techniques is critical to understanding their structure-function relationship. Microstructural images, alongside statistical image analysis and physical simulations, allow for microstructural features to be linked to material behaviour. This in turns facilitates material design and optimisation. Unfortunately, common imaging techniques can suffer from unwanted artefacts during sample preparation, or interference and disturbances during imaging, resulting in regions of the image being unusable for analysis and simulation (hereafter referred to as occluded regions). Some examples of these are surface scratches during sample cutting, charging effects in scanning electron microscopy (SEM) and regions of an image being corrupted by software errors. Whilst the characterisation procedure could be repeated in the hope of obtaining a clean micrograph (if the imaging technique is not destructive), this is a time consuming process, and particularly challenging if large, representative images are needed. Alternatively, it is possible to replace these occluded regions with reconstructed microstructure post-hoc, saving time, money and sample. This process is called inpainting. 

Broadly, there are two approaches to inpainting - classical statistical reconstruction and machine learning reconstruction. The most ubiquitous statistical reconstruction technique is exemplar-based inpainting, whereby the occluded region is filled in from the outer edge to center with the best matching patches that are copied-and-pasted from the unoccluded region \cite{Liu2013-cd}. Barnes \textit{et al.} proposed the PatchMatch algorithm for fast patch search using the natural coherency of the image \cite{Barnes2009-oh}. Tran \textit{et al.} extended the PatchMatch algorithm to microstructural inpainting \cite{Tran2021-lv}. Their method aims to compliment machine learning approaches that typically require large datasets. This approach is successful in reconstructing grayscale, experiment level data, however, the reconstructed region can contain exactly copied features which is unrealistic and potentially unrepresentative.

Convolutional neural networks (CNNs) form the basis of many visual machine learning tasks. The majority of generative methods that use CNNs take the form of autoencoders, diffusion models or generative adversarial networks (GANs). General purpose inpainting models using these methods have been developed, which have been extremely successful across many applications \cite{Pathak2016-uj, Yan2018-ln, Demir2018-rf, Lugmayr2022-jx}. However, many of the state-of-the-art (SOTA) models require large labelled datasets for training from scratch or fine-tuning pretrained models. This makes it near impossible to apply such techniques in the field of material characterisation, where data availability is severely limited and time consuming to collect. SOTA models are also often very deep to allow the synthesis of a wide variety of complex features, which leads to long,  memory and compute intensive training. Such techniques have poor accessibility to the general community due to the expensive computational resources that are required. This collection of unique requirements for microstructure inpainting, including limited novel data availability and the desire for fast, low compute training, reveals the opportunity for specific inpainting methods for material science.

Microstructural image data has properties that can be exploited in order to address the issues outlined above. Often micrographs are taken of the bulk of a material, and the resulting data is homogeneous, isotropic and representative. Therefore, any large enough patch of microstructure is statistically equivalent to any other patch. This allows a single image to be batched into a statistically equivalent distribution of smaller images, hence forming a training dataset for a generation algorithm. This eases the requirement on collecting a large dataset from many images. An additional simplification that derives from the isotropy and homogeneity of the data, is that if the statistical distribution has been learned by a generative model, the spatial dimensions of the generated output can be adapted arbitrarily, allowing for beyond lab capability generation of synthetic data. This idea was demonstrated by Gayon-Lombardo \textit{et al.} using an adjustable latent vector in a GAN framework \cite{Gayon-Lombardo2020-mr}. The required output size of the network must match the largest features of the dataset, not the size of the entire image. This means networks can be much smaller than for fixed output size problems. The relative simplicity of microstructural features also reduces the required number of parameters, shrinking training times and reducing memory requirements. These properties of microstructural data greatly reduce the memory and compute requirements when training generation algorithms, and also mitigate the need for large training datasets. However, it also means the data must conform to the assumptions of isotropy, homogeneity and representativity.

GANs are a family of machine learning models characterised by the use of two networks competing in an adversarial game. They are capable of generating samples from an underlying probability distribution of an input training dataset. Mosser \textit{et al.} introduced GANs as a method for reconstructing synthetic realisations of a homogeneous microstructure \cite{Mosser2017-tz}. Further methods have been developed to reconstruct 3D multi-phase microstructure, and generate 3D images from 2D data \cite{Gayon-Lombardo2020-mr, Kench2021-rm}. These models make use of some assumptions about microstructural image data outlined earlier to shrink the memory and compute requirements of training. GANs have emerged as the most common machine learning method for inpainting microstructure. Ma \textit{et al.} developed an automatic inpainting algorithm which involves two steps, firstly the classification and segmentation of the occluded region, followed by inpainting \cite{Ma2021-lf}. A U-Net performs the segmentation of the damaged region, and an EdgeConnect model performs the inpainting \cite{Ronneberger2015-vn, Nazeri2019-cp}. This method requires a large dataset of manually labelled damaged regions, which makes this method hard to generalise to all types of defect. Karamov \textit{et al.} developed a GAN-based method, with an autoencoder generator for inpainting grayscale, 3D, anisotropic micro-CT images \cite{Karamov2021-xq}. This method demonstrated moderate success but struggled to form contiguous boundaries. The resulting inpainted microstructure had an observable hard border of non-matching pixels.

This paper outlines two novel GAN-based methods for inpainting microstructural image data without the need for large datasets or labelled data. These methods are designed to be applied in different scenarios. Each approach seeks to satisfy two key requirements for successful inpainting, namely the generation of realistic features to replace the occluded region, and the matching of these features to existing microstructure at the inpainting boundary. The first method, generator optimisation (G-opt), uses a combination of a standard GAN loss (maximise realness of generated data) and a content loss (minimise difference between generated and ground truth boundary) to simultaneously address both goals. The resulting generator is well optimised for a specific inpaint region, but cannot be applied to other defect regions. The second method, seed optimisation (z-opt), decouples the two requirements by first training a GAN to generate realistic microstructure, and then searching the latent space for a good boundary match. This means the generator can be applied to any occluded region after training, but boundary matching can be less successful. It is important to note that these methods are stochastic, and that the inpainted region is not meant to reconstruct the ground truth. Instead these techniques aim to synthesise entirely new, unseen regions of microstructure, whilst maintaining a contiguous border with the unoccluded region. Due to the stochastic nature of these methods, there is no single solution to this problem, and a family of solutions can be synthesised.

Additionally, in section \ref{sec:GUI} this paper presents a graphical user interface (GUI) through which users can easily apply these methods to their own data. The purpose of this is to provide democratic access to a tool for materials scientists from a range of disciplines. The GUI requires little to no coding experience and has been made open source.

\section{Methods}
\label{sec:methods}

In this section the generator optimisation method and then the seed optimisation method will be described. Following this, the inpainting quality analysis methods will be introduced.

\subsection{Generator optimisation}

As previously described, the G-opt technique involves training a generator model that can synthesise realistic inpainted regions. It incorporates a content loss between the generated and real boundary to enforce feature continuity, and a conventional Wasserstein loss to ensure realistic features \cite{Arjovsky2017-vr, Gulrajani2017-bt}. Algorithm \ref{alg:GOptAlgo} describes the training regime for G. When computing the content loss, a fixed seed is used as the input, then the MSE is calculated on an annulus of the real and fake data. The fixed seed is kept the same throughout training when calculating the content loss, and saved alongside the model weights after training, as it is this specific seed which G learns to map to the matching boundaries. It is important to note that during training, the Wasserstein loss must be calculated without the fixed seed region, and is instead trained with a random seed, as otherwise the constant annulus that it generates could be used by the discriminator to identify fake samples. 

\begin{algorithm*}[htb!] 
	\begin{algorithmic}[1]
	\Require $G$, the generator function; $D$, the discriminator function; $c$, the coefficient of the content loss multiplication (default $1$); $gt$, the ground truth region selected for inpainting; $CL$, the content loss function (mean squared error); $AN$, annulus function that takes only the pixels in the 16 pixel boundary around the occluded region; $S$, seed function that takes the fixed seed and replaces possible central elements with random values; $gp$, gradient penalty; $i_{max}$, maximum number of iterations; $\mathcal{N}(0,1)$, standard normal distribution; All batch operations and optimization parameters are not shown for simplicity, we refer the reader to the codes at the project's repository for specific parameter details, including details about the implementation of the gradient penalty.

    \Statex
	\Statex\(\triangleright\){\% Training \%}
	    \State select inpaint region of size $d \times d$
	    \State $z_{fixed}\leftarrow z \sim\mathcal{N}(0,1)$ sample a fixed seed of size $s \times s$, $s = \lfloor\frac{d}{8}\rfloor + 6$
		\For {$i = 0, ..., i_{max}$}
		\State $z_{rand}\leftarrow z\sim\mathcal{N}(0,1)$
		\State $fake \leftarrow G(z_{rand})$
		\State $real\leftarrow$ sample a batch of training images from the unoccluded region.
		\State $l_{D} \leftarrow D(fake) - D(real) + gp$
		\State backpropagate and update the weights of $D$ from the loss $l_{D}$.
		\If {$10 \mid i$}
		\State $z_{rand}\leftarrow z\sim\mathcal{N}(0,1)$
		\State $fake \leftarrow G(z_{rand})$
		\State $fixed \leftarrow G(z_{fixed})$ output has dimensions $(d+16) \times (d+16)$
		\State $l_{CL} \leftarrow CL(AN(gt), AN(fixed))$ take content loss between the $16$ pixel annulus of $gt$ and $fixed$
		\State $l_{G} \leftarrow -D(fake) + c\cdot l_{CL}$
	    \State backpropagate and update the weights of $G$ from the loss $l_{G}$.
		\EndIf
		\EndFor
	\Statex
	\Statex\(\triangleright\){\% Evaluation \%}
	    \State {$z_{eval} \leftarrow S(z_{fixed})$}
	    \State {$out \leftarrow G(z_{eval})$}
	    \State {$inpaint \leftarrow out, gt$} Replace $d\times d$ occluded region with output of G (with annulus removed)
		\caption{{\footnotesize Generator optimisation algorithm}}
		\label{alg:GOptAlgo}
	\end{algorithmic}
\end{algorithm*}

Interestingly, the architecture of G remains the same for all sizes of occluded region. To match the network output size to the occluded region, we change the spatial size of the random input seed, which is an established technique for controlling image dimensions when generating homogeneous textures of material micrographs. In the standard network that we use, increasing the input seed size by 1 results in an output size increase of 8. The size of the selected occluded region is thus restricted to be a multiple of 8 pixels in each dimension, allowing for an associated integer seed size. This calculated seed size is increased by two in each direction when passed to G in order to generate the boundary region on which the content loss is calculated. 

During evaluation, the fixed seed is passed to G and the boundary region of the output is replaced with the boundary region of the original image, such that only the occluded region is replaced. For occluded regions larger than $64\times64$ pixels, a central part of the seed can be replaced with random noise to create a stochastically varying inpainted region with fixed boundaries. It is important not to change seeds too close to the boundary, as the propagation of information through the generator with the transpose convolutions means these seeds will influence the boundary region, and must thus remain fixed. Further description of how to determine which seeds can be safely changed is given in the Supplementary Information. 


\subsection{Seed optimisation}

The z-opt approach separates the tasks of generating realistic microstructure and generating well matched contiguous boundaries. The generator is trained with the usual Wasserstein loss metric and the seed is optimised for inpainting after training. The decoupling of the two optimisation tasks enables the generator to inpaint any region of the micrograph after it has been trained.

The z-opt is performed by first calculating the MSE between the annulus of the generated region and the ground truth. Then, whilst holding the weights of G constant, the MSE is backpropagated to the seed, which is treated as a learnable parameter. If the iterative updates to the seed are unconstrained, its distribution of values deviate significantly from the random normal noise distribution used during training. This is problematic, as although the resulting MSE on the boundary is potentially very low, the central features in the occluded region become unrealistic. 

Initially, we attempt to address this deviation through a simple re-normalisation procedure of the seed after each seed update, which can be implemented by subtracting the mean of the seed and dividing by the standard deviation. However, the output of the optimisation after many iterations appeared to deviate from realistic microstructure and become blurry. A histogram of the values of the seed showed that the seed became non-normally distributed, and although retained a mean of 0 and standard deviation of 1, it in fact became bimodal, with two peaks centered around 1 and -1. To keep the seed normally distributed, a KL loss (a statistical measure of distance between two distributions) between the seed and a random normal seed was introduced which anchored the optimised seed to the distribution of random normal seeds. This stopped the more unrealistic features being generated and enforced a normal distribution throughout the optimsation process.

The exact architecture for G and D can be tweaked for different use cases. To fairly compare the methods, it was decided that the same network architectures and hyperparameters should be used for both methods. These are shown in the Supplementary Information Table \ref{tab:networks}. In this study, we explore inpainting of n-phase, colour and grayscale images. For n-phase segmented images, the final activation layer of the generator is an n-channel softmax, where the output value in each channel corresponds to the `confidence' that the given pixel belongs to the phase. In post-processing, the maximum confidence phase is selected and the output is shown as segmented into three phases. Otherwise, the final layer of the output is a sigmoid. In the results presented in this paper, both methods are also trained for the same number of iterations and for each method no cherry picking of results was performed.

\subsection{Inpaint quality analysis}
\label{sec:validation}
There are two important aspects to validating the `goodness' of the inpainting. Firstly, the inpainted microstructure must retain the statistics of the training data. Secondly, the border between the generated microstructure and original data must be well matched. To validate the first of these, the distribution of volume fractions (VFs) of generated microstructure was calculated. To analyse the contiguity of the border, a technique was developed that compares the distribution of mean squared errors between neighbouring pixels on the boundary. 

\begin{figure}[!h]
\centering
    {\includegraphics[width=0.75\textwidth]{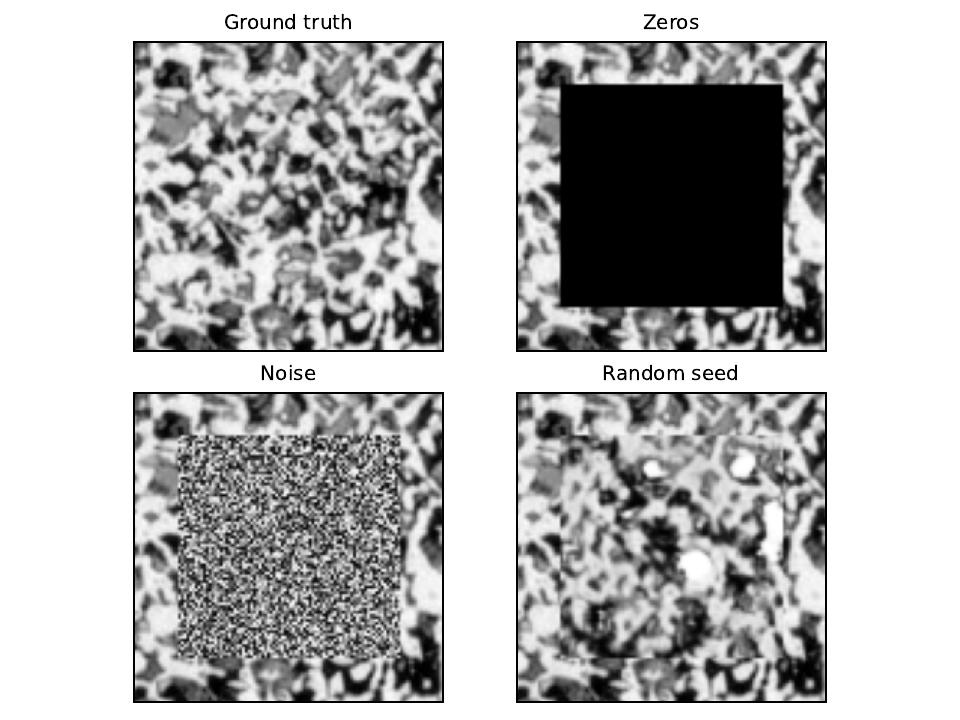}}
    {\caption{The four inpainting examples used to test the border contiguity analysis. The p-values for each test: $\text{GT} = 0.017,\ \text{Zeros} = 1.7\times10^{-184},\ \text{Noise} = 8.2\times10^{-83},\ \text{Random seed} = 6.1\times10^{-46}$.}
    \label{fig:border}}
\end{figure}

First, we calculate the difference between the pixels on the outside edge of the inpainted region (which belong to the original image) and the pixels on the inside edge of the inpainted region (which belong to the generated image). The square of these differences form a distribution that describes the mean squared error of neighbouring pixels. A ground truth distribution is then calculated by taking the mean squared error between all neighbouring pixels in the original image. A KS test is then used to return the probability that the distribution calculated from the inpainted border and the distribution calculated from the ground truth are the same. For comparison, this border contiguity test was also performed on an inpainting of zeros, uniform noise and the output of a trained generator given an unoptimised random input seed as shown in Figure \ref{fig:border}.

The p-value for the ground truth gives a reference value for what `perfect' inpainting looks like for this microstructure, and the order of magnitude of the p-value can be used to compare different inpainting methods, and quantify how discontiguous the border of the inpainting is relative to the ground truth. The ground truth p-value is not necessarily 1, as the KS test is performed between the MSE distributions of neighbouring pixels across the whole image and the border of the `to be' inpainted region. We expect the p-value to be closer to one the more the border region is representative of the global distribution.

\section{Results}
\label{sec:results}

\subsection{N-phase}

\begin{figure}[!h]
\centering
    {\includegraphics[width=0.75\textwidth]{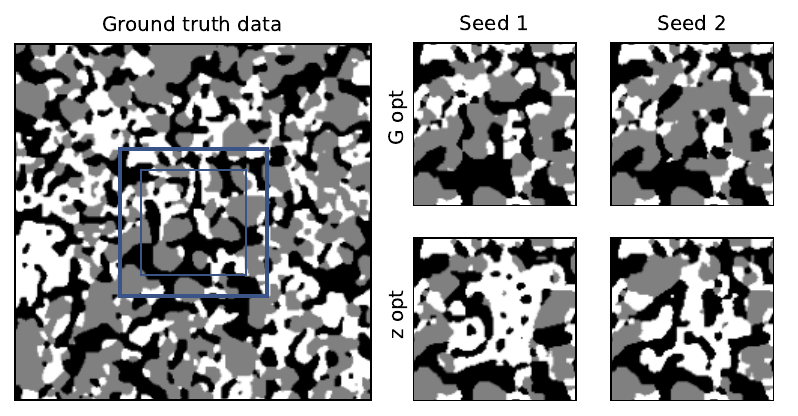}}
    {\caption{The inpainting result for the two methods on a three-phase dataset. The p-values for the contiguity analysis: $\text{GT} = 1,\ \text{G-opt} = 1,\ \text{z-opt} = 1,\ \text{zeros} = 2.2 \times 10^{-59},\ \text{noise} = 3.1\times 10^{-120},\ \text{random seed} = 6 \times 10^{-192}$.}
    \label{fig:case1}}
\end{figure}

Initially, both methods were tested on a three-phase solid oxide fuel cell anode \cite{Hsu2018-az}. The material was imaged using FIB-SEM and then segmented into three phases: pore (black), metal (grey) and ceramic (white). An occluded region was chosen that did not contain any defects for the purpose of comparison between the ground truth and the generated output. The results of the inpainting with both methods is shown in Figure \ref{fig:case1}.

\begin{figure}[!h]
\centering
    {\includegraphics[width=0.75\textwidth]{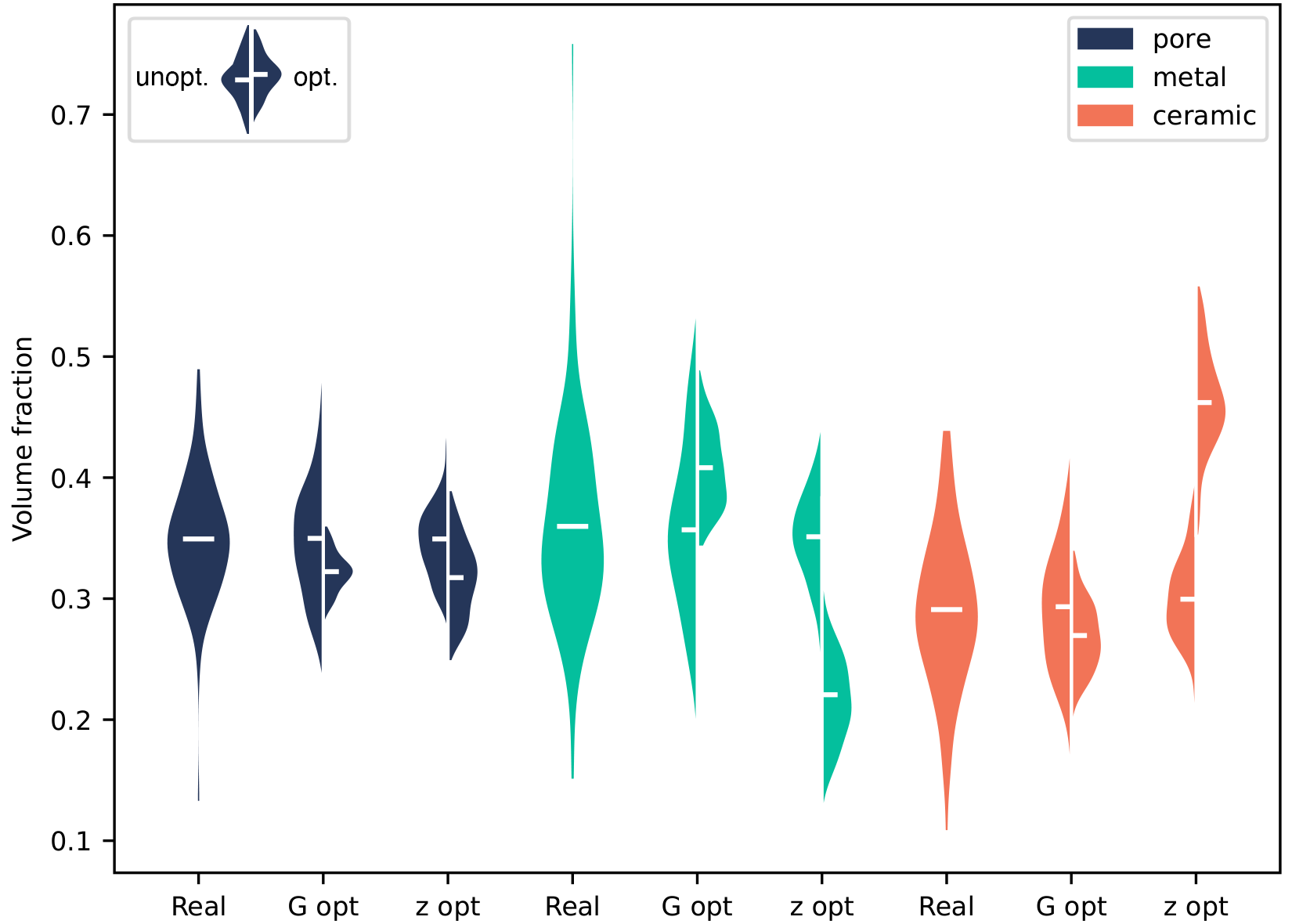}}
    {\caption{Volume fraction distributions for the three phases of case 1 - SOFC anode. For each phase, the distribution of volume fractions for the ground truth data (Real) is shown. The G-opt is split into half violins for a random input seed (left) and for the hybrid fixed-random input seed (right). The z-opt is split into half violins for a random input seed (left) and an optimised input seed (right). The distributions are shown across 128 different seeds, but all on the same inpainted region. The mean of the distribution is shown as a white bar.}
    \label{fig:vfs}}
\end{figure}

Figure \ref{fig:vfs} shows the results of volume fraction analysis on the inpainted microstructures. By enforcing the boundary of our generated volume to match, we naturally restrict the space of possible structures, and therefore we do not expect to exactly recover the same VF distribution as the ground truth data. However, we do expect our generator to be capable of producing this distribution when given a random, unoptimised seed. Therefore, in Figure \ref{fig:vfs}, for each method, the unoptimised and optimised cases are compared.

We first consider the unoptimised case. KS tests were performed on each method to compare the distributions of volume fractions to the ground truth, the full results are shown in the Supplementary Information. The p-value is a measure of how probable it is these samples were taken from the same distribution. With a random seed as input, the G-opt method produces distributions with large p-values (0.73-0.97), indicating a good agreement with the ground truth distribution. A random seed given to the z-opt method produces smaller p-values (0.022-0.43), revealing poorer agreement with the ground truth. As these generators are identical in architecture and were trained for the same number of iterations, this indicates that the addition of the content loss during training improves the overall quality of the generator. 

It is possible that because the content loss is introduced from the start of training, G can immediately start to learn kernels that produce realistic features, without requiring useful information from D. This inevitably speeds up the convergence of G, and also aids in training D, as `realness' of the output of G will be improved earlier in training. Without this content loss, G is entirely reliant on the information from D, and therefore cannot start to learn realistic features until D has learned to start distinguishing these.

Figure \ref{fig:vfs} shows that the optimised seed VF distributions of the G-opt method, are constrained within the bounds of the distribution produced from the random seed input. This suggests that, although the VF distribution of the optimised seed is not similar to the ground truth distribution, the VFs of the generated microstructure are at least a subset of the underlying VF distribution. On the other hand, the z-opt distribution with an optimised seed is significantly offset from the distribution produced by random input seeds. This can be explained by the z-opt process. During training, G is given seeds that are sampled from a random normal distribution. When the seed is optimised post-training, the optimisation pushes the seed into a region where the boundary is best matched, and although the seed is encouraged to retain its normality, this region of latent space may not have been well sampled in training, therefore generating samples that do not follow the same statistics as the underlying data.

To quantify the contiguity of the border, the analysis outlined in Section \ref{sec:validation} was performed on the inpainted result of both methods. This analysis reveals the G-opt method produces borders that are indistinguishable from the ground truth, yielding a p-value of 1. The z-opt method performs worse, and produces a more significant result, despite the border not being noticeably discontiguous.

\subsection{Grayscale and colour}

The second case presented in this study is a grayscale image of a hypoeutectoid steel (micrograph 237) taken from DoITPoMS \cite{}. Obtaining a realistic and contiguous output is more difficult for continuous pixel values, and therefore the networks were trained for twice as long (200k iterations). 

\begin{figure}[!h]
\centering
    {\includegraphics[width=0.75\textwidth]{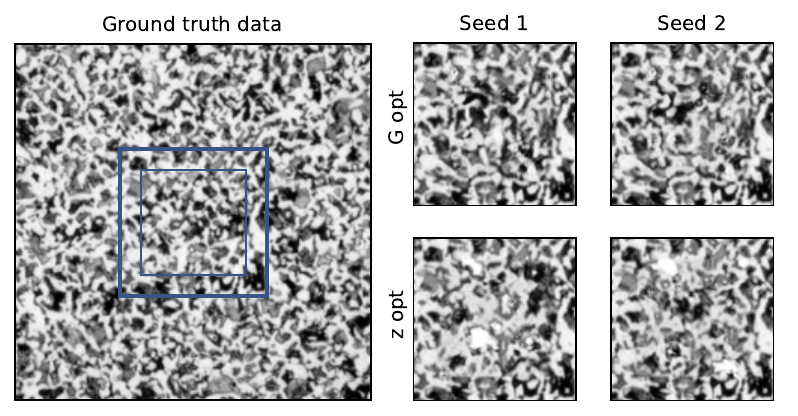}}
    {\caption{The inpainting result for the two methods on a grayscale dataset. The p-values for the contiguity analysis: $\text{GT} = 0.017,\ \text{G-opt} = 1.4\times10^{-6},\ \text{z-opt} = 2.7\times10^{-14},\ \text{zeros}=1.7\times 10^{-184},\ \text{noise} = 8.2\times 10^{-83},\ \text{random seed} = 6.1\times 10^{-46}$.}
    \label{fig:case2}}
\end{figure}

Analysis of volume fraction of phases is not possible for unsegmented data, which makes assessing the quality of the generated output challenging. It is possible to compare pixel value distributions, and we refer the reader to the Supplementary Information for more details. In this section we will focus on the contiguity analysis for grayscale and colour, and make the assumption that the behaviours observed in the n-phase case are retained. The results reveal a more noticeable disparity in the contiguity analysis between the ground truth, G-opt method and z-opt method. The generator method outperforms the z-opt method by an order of magnitude, but is now significantly lower than the ground truth. Inspecting the inpainting visually, small discontiguities in the G-opt method are visible. The z-opt method shows clearly visible boundaries, and also some unrealistic features emerging.


The third case is a colour image of a terracotta pot (micrograph 177) taken from DoITPoMS \cite{}. As colour is an additional level of complexity, the model was trained for 300k iterations. For this case, the occluded region contains a material artefact. Contiguity analysis reveals a stark difference in the performance of the two methods, also corroborated by visual inspection. The G-opt method outperforms the z-opt method by many orders of magnitude, and the borders appears much more contiguous.

\begin{figure}[!h]
\centering
    {\includegraphics[width=0.75\textwidth]{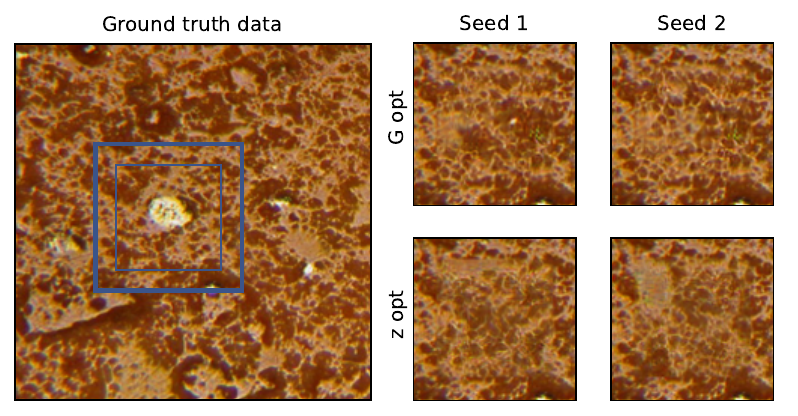}}
    {\caption{The inpainting result for the two methods on a colour dataset. The p-values for the contiguity analysis: $\text{GT} = 4.7\times10^{-5},\ \text{G-opt} = 1.1\times10^{-13},\ \text{z-opt} = 3.7\times10^{-46},\ \text{zeros} = 0,\ \text{noise} = 0,\ \text{random seed} = 6.5\times 10^{-170}$.}
    \label{fig:case3}}
\end{figure}


\section{Discussion}

Both methods perform better for n-phase, segmented images. The generated inpainting will necessarily match the boundaries better, as the final output of the generator is a softmax which means that high probability predictions result in exact matching of phases after segmentation. A marked difference in the accuracy of the two methods arises in the application to grayscale and colour images. Grayscale and colour inpainting require the outputs of the generator to be continuous values, and therefore the boundary matching problem is inherently harder. In these cases, the G-opt method outperforms the z-opt method. Intuitively, we can understand this disparity by considering the conditioning of the latent space during training. When the output space becomes continuous rather than discrete, the space of possible microstructures is larger. Therefore, it becomes even less likely that the seed that corresponds to a well matched boundary will exist in a well-conditioned region of the latent space without constraining the space. The G-opt method introduces this constraint, and ensures the seed exists in the latent space. 

Variation in the inpainted region when changing the random seed implies over-fitting has not occured during training. This demonstrates that the proposed methods do not require large datasets for training. They do, however, rely on the assumption that the data is homogeneous and that the unoccluded region is statistically representative of the material.

\begin{figure}[!h]
    {\includegraphics[width=\textwidth]{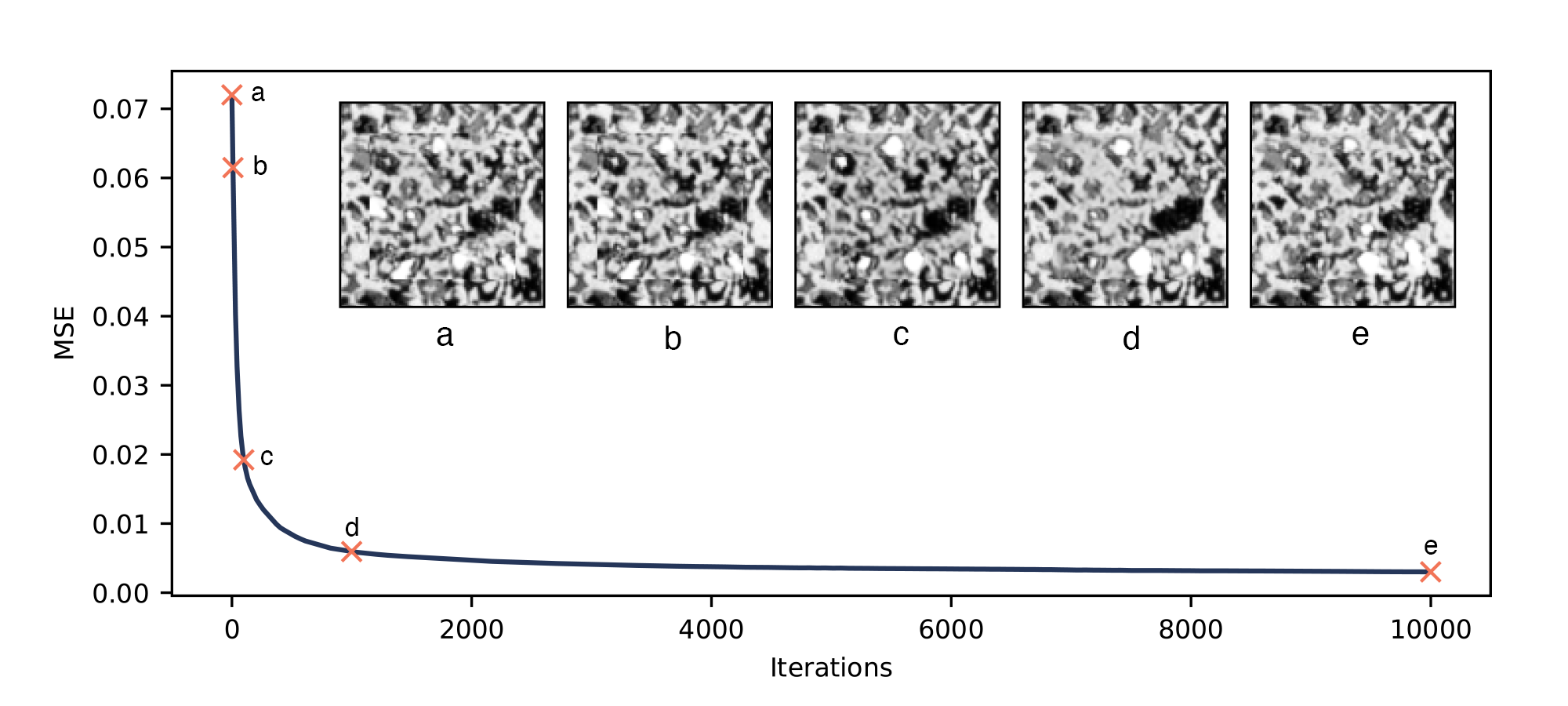}}
    {\caption{The output of G and the MSE throughout the seed optimisation process.}
    \label{fig:z_span}}
\end{figure}

The optimisation of the seed to minimise the content loss appears to push the generator to generate unrealistic microstructure. This was confirmed by the distribution of VFs in Figure \ref{fig:vfs}. Figure \ref{fig:z_span} shows the inpainted micrograph during seed optimisation. This demonstrates that as the seed is optimised the boundary becomes better matched, but some unrealistic features emerge. It is interesting to note that the intermediate results after 100 and 1000 iterations are particularly unrealistic, and that the microstructure returns to more realistic at long optimisation times. This could be because the seed first seeks to satisfy the easier MSE condition on the border, and then searches for a more normal seed distribution to satisfy the KL loss. It is clear that the seed that corresponds to a perfect matching boundary either does not lie in the space of realistic microstructures or at least this process is unable to satisfy both conditions, hence motivating the alternate method.

As previously mentioned, both methods were trained using the same hyperparameters, with the only difference in the training procedure being the fixing of the seed and the inclusion of the content loss. The G-opt method therefore takes longer per iteration. However, once trained the G-opt method is much faster to evaluate, with the z-opt requiring a new optimisation for each new instance. Overall, there is a trade-off between training time, generation time and quality, meaning a method should be chosen according to the application.

Ultimately, the user determines whether or not the model or optimisation has converged. The hyperparameters in this paper are a guide, but can be tuned for different use cases. For example, for more complex materials, the number of filter layers in the networks can be increased, the training time extended and the number of optimisation iterations increased. The volume fraction and border contiguity analysis outlined in this paper are useful guides when comparing different methods and sets of hyperparameters. However, a universal, quantitative metric was not found to measure convergence across all materials, and therefore the user is encouraged to judge convergence by visual inspection. 

\section{Graphical User Interface}
\label{sec:GUI}

Inpainting is a very visual problem, involving multiple steps that require visual feedback, from the identification of occlusions, to the evaluation of performance. Therefore, a graphical user interface (GUI) was developed alongside the command line interface to support a more visual workflow, as well as enabling users with less coding experience. 

\begin{figure}[!h]
    {\includegraphics[width=\textwidth]{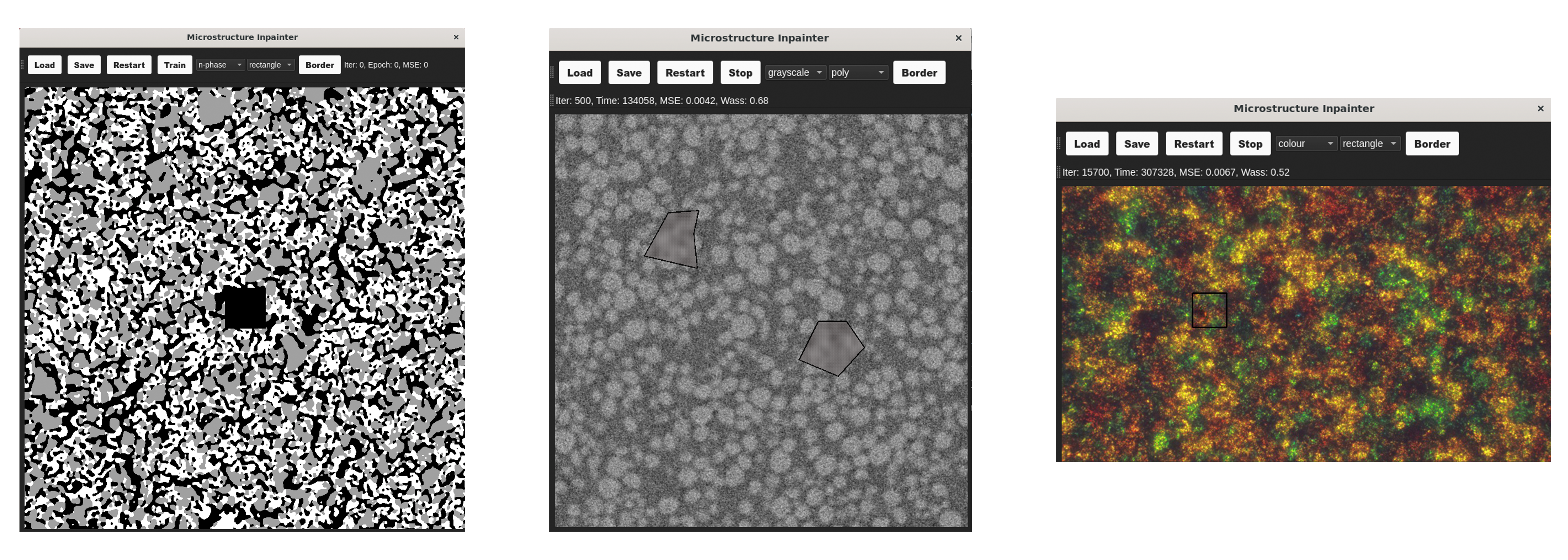}}
    {\caption{The graphical user interface.}
    \label{fig:gui}}
\end{figure}

The GUI is designed for quick and simple use of the tool. The user flow is roughly as follows: 
\begin{enumerate}
    \item Loads in an image to inpaint from their files.
    \item Selects the image type and desired method.
    \item Draws either a rectangle or polygon around the occluded regions.
    \item Initiates training.
    \item Watches as the image is updated with the models attempt at inpainting during training.
    \item Decides if the model has converged and stops training. 
    \item Generates new instances of the inpainted region.
    \item Saves the inpainted image as a new file.
\end{enumerate}

At present, the rectangle drawing shape has been implemented for the G-opt method and the polygon drawing method for the z-opt method. This is due to the relative ease of implementation. However, there is no fundamental reason why the two methods could not be adapted in the future to solve for the alternate shape types. Additional further work on the GUI will include a saving and loading models option, threading the optimisation of the seed during training for speed and an option to edit the hyperparameters and model architecture via the GUI. For the time being, the GUI can be built locally, allowing the user to adjust the finer details of the method. If this is not required, the GUI can be run from a downloadable executable file, requiring no coding experience or knowledge.

This work can be trivially extended to 3D inpainting. The extension to 3D microstructural GANs has been demonstrated in multiple applications \cite{Kench2021-rm, Gayon-Lombardo2020-mr}. All that would be required would be to replace 2D (transpose-) convolutions with 3D (transpose-) convolutions, and add a spatial dimension to the seed. The difficult aspect of extending to 3D would be identifying 3D defects through a simple visual interface, which is beyond the scope of this work.

\section{Conclusion}

Two complementary inpainting methods have been developed using deep convolutional generative adversarial networks. The two methods have relative merits, but overall the generator optimisation method outperforms the seed optimisation method on two important measures of realism of generated output and contiguity of borders. Both methods performed more strongly for n-phase, segmented images than colour or grayscale. Visual comparison to existing microstructural inpainting methods indicates improved border contiguity. Additionally, the methods can be applied via a command line interface or graphical user interface which has been made free and open source, allowing access to users with no coding experience. These two methods offer a fast and convenient way of inpainting microstructural image data, and will hopefully lead to images with a range of defects being made usable for characterisation and modelling.


\section*{Carbon Emissions}

Experiments were conducted using our own workstation in London, which has a carbon efficiency of approximately 0.193 kgCO$_2$eq/kWh. A cumulative of 275 hours of computation was performed on hardware of type NVIDIA RTX A6000 (TDP of W).

Total emissions are estimated to be 15.92 kgCO$_2$eq. Estimations were conducted using the \href{https://mlco2.github.io/impact/}{Machine Learning Emissions calculator} presented in \cite{Lacoste2019-jn}. In order to contain all other emissions from the use of personal computers, commuting etc. A reported 1 tonneCO$_2$eq of carbon offset was bought from \href{https://native.eco/}{Native Energy}.

\section*{Funding Statement}
SJC and SK were supported by funding from the EPSRC Faraday Institution Multi-Scale Modelling project (https://faraday.ac.uk/; EP/S003053/1, grant number FIRG003 received by AD, SK and SJC), and funding from the President’s PhD Scholarships received by AD. Funding from the Henry Royce Institute's Materials 4.0 initiative was also received by AD and SJC (EP/W032279/1).

\section*{Competing Interests}
None

\section*{Data Availability Statement}
All data used in this study is publicly available through the references provided, and in the GitHub repository for the project.

\section*{Code Availability}
The codes used in this manuscript are available at \url{https://github.com/tldr-group/microstructure-inpainter} with an MIT license agreement.

\section*{Ethical Standards}
The research meets all ethical guidelines, including adherence to the legal requirements of the study country.

\section*{Author Contributions}
Conceptualization: I.S; S.K. Methodology: I.S; S.K. Writing original draft: I.S. All authors approved the final submitted draft.

\section*{References}
\addcontentsline{toc}{section}{References}
\def\addvspace#1{}

	\renewcommand{\refname}{ \vspace{-\baselineskip}\vspace{-1.1mm} }
	\bibliographystyle{unsrt}
    \bibliography{paperpile}

\begin{thebibliography}{10}

\bibitem{Liu2013-cd}
Yunqiang Liu and Vicent Caselles.
\newblock Exemplar-based image inpainting using multiscale graph cuts.
\newblock {\em IEEE Trans. Image Process.}, 22(5):1699--1711, May 2013.

\bibitem{Barnes2009-oh}
Connelly Barnes, Eli Shechtman, Adam Finkelstein, and Dan~B Goldman.
\newblock {PatchMatch}: A randomized correspondence algorithm for structural
  image editing.
\newblock {\em ACM Trans. Graph.}, 28(3), August 2009.

\bibitem{Tran2021-lv}
Anh Tran and Hoang Tran.
\newblock {2D} microstructure reconstruction for {SEM} via non-local
  {Patch-Based} image inpainting.
\newblock In {\em {TMS} 2021 150th Annual Meeting \& Exhibition Supplemental
  Proceedings}, pages 495--506. Springer International Publishing, 2021.

\bibitem{Pathak2016-uj}
Deepak Pathak, Philipp Krahenbuhl, Jeff Donahue, Trevor Darrell, and Alexei~A
  Efros.
\newblock Context encoders: Feature learning by inpainting.
\newblock April 2016.

\bibitem{Yan2018-ln}
Zhaoyi Yan, Xiaoming Li, Mu~Li, Wangmeng Zuo, and Shiguang Shan.
\newblock {Shift-Net}: Image inpainting via deep feature rearrangement.
\newblock January 2018.

\bibitem{Demir2018-rf}
Ugur Demir and Gozde Unal.
\newblock {Patch-Based} image inpainting with generative adversarial networks.
\newblock March 2018.

\bibitem{Lugmayr2022-jx}
Andreas Lugmayr, Martin Danelljan, Andres Romero, Fisher Yu, Radu Timofte, and
  Luc Van~Gool.
\newblock {RePaint}: Inpainting using denoising diffusion probabilistic models.
\newblock January 2022.

\bibitem{Gayon-Lombardo2020-mr}
Andrea Gayon-Lombardo, Lukas Mosser, Nigel~P Brandon, and Samuel~J Cooper.
\newblock Pores for thought: generative adversarial networks for stochastic
  reconstruction of {3D} multi-phase electrode microstructures with periodic
  boundaries.
\newblock {\em npj Computational Materials}, 6(1), 2020.

\bibitem{Mosser2017-tz}
Lukas Mosser, Olivier Dubrule, and Martin~J Blunt.
\newblock Reconstruction of three-dimensional porous media using generative
  adversarial neural networks.
\newblock {\em Phys Rev E}, 96(4-1):043309, October 2017.

\bibitem{Kench2021-rm}
Steve Kench and Samuel~J Cooper.
\newblock Generating three-dimensional structures from a two-dimensional slice
  with generative adversarial network-based dimensionality expansion.
\newblock {\em Nature Machine Intelligence 2021 3:4}, 3(4):299--305, April
  2021.

\bibitem{Ma2021-lf}
Boyuan Ma, Bin Ma, Mingfei Gao, Zixuan Wang, Xiaojuan Ban, Haiyou Huang, and
  Weiheng Wu.
\newblock Deep learning-based automatic inpainting for material microscopic
  images.
\newblock {\em J. Microsc.}, 281(3):177--189, March 2021.

\bibitem{Ronneberger2015-vn}
Olaf Ronneberger, Philipp Fischer, and Thomas Brox.
\newblock {U-Net}: Convolutional networks for biomedical image segmentation.
\newblock May 2015.

\bibitem{Nazeri2019-cp}
Kamyar Nazeri, Eric Ng, Tony Joseph, Faisal~Z Qureshi, and Mehran Ebrahimi.
\newblock {EdgeConnect}: Generative image inpainting with adversarial edge
  learning.
\newblock January 2019.

\bibitem{Karamov2021-xq}
Radmir Karamov, Stepan~V Lomov, Ivan Sergeichev, Yentl Swolfs, and Iskander
  Akhatov.
\newblock Inpainting {micro-CT} images of fibrous materials using deep
  learning.
\newblock {\em Comput. Mater. Sci.}, 197:110551, September 2021.

\bibitem{Arjovsky2017-vr}
Martin Arjovsky, Soumith Chintala, and L{\'e}on Bottou.
\newblock {W}asserstein generative adversarial networks.
\newblock In Doina Precup and Yee~Whye Teh, editors, {\em Proceedings of the
  34th International Conference on Machine Learning}, volume~70 of {\em
  Proceedings of Machine Learning Research}, pages 214--223. PMLR, 2017.

\bibitem{Gulrajani2017-bt}
Ishaan Gulrajani, Faruk Ahmed, Martin Arjovsky, Vincent Dumoulin, and Aaron
  Courville.
\newblock Improved training of wasserstein {GANs}.
\newblock March 2017.

\bibitem{Hsu2018-az}
Tim Hsu, William~K Epting, Rubayyat Mahbub, Noel~T Nuhfer, Sudip Bhattacharya,
  Yinkai Lei, Herbert~M Miller, Paul~R Ohodnicki, Kirk~R Gerdes, Harry~W
  Abernathy, Gregory~A Hackett, Anthony~D Rollett, Marc De~Graef, Shawn
  Litster, and Paul~A Salvador.
\newblock Mesoscale characterization of local property distributions in
  heterogeneous electrodes.
\newblock {\em J. Power Sources}, 386:1--9, May 2018.

\bibitem{Lacoste2019-jn}
Alexandre Lacoste, Alexandra Luccioni, Victor Schmidt, and Thomas Dandres.
\newblock Quantifying the carbon emissions of machine learning.
\newblock October 2019.

\end{thebibliography}

\section*{Supplementary Information}

\subsection*{Training data}
\begin{figure}[H]
\centering
\begin{subfigure}{0.4\textwidth}
    \centering
    {\includegraphics[width=\textwidth]{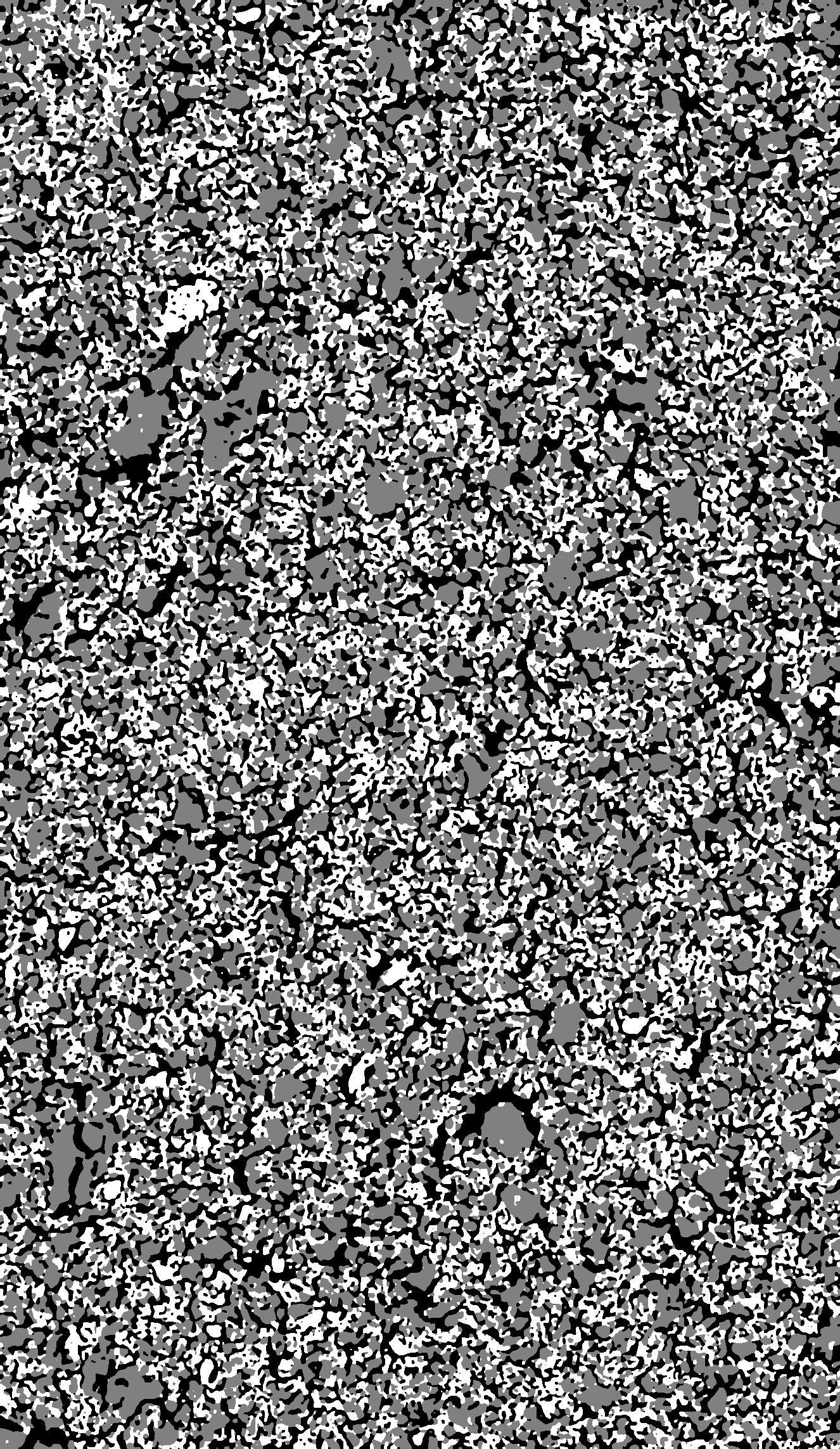}}
    {\caption{Case 1 - 3-phase segmented SOFC micrograph \cite{Hsu2018-az}. $1326 \times 2286$ pixels.}
    \label{fig:data1}}
    \end{subfigure}
\begin{subfigure}{0.4\textwidth}
\begin{subfigure}{\textwidth}
    \centering
    {\includegraphics[width=\textwidth]{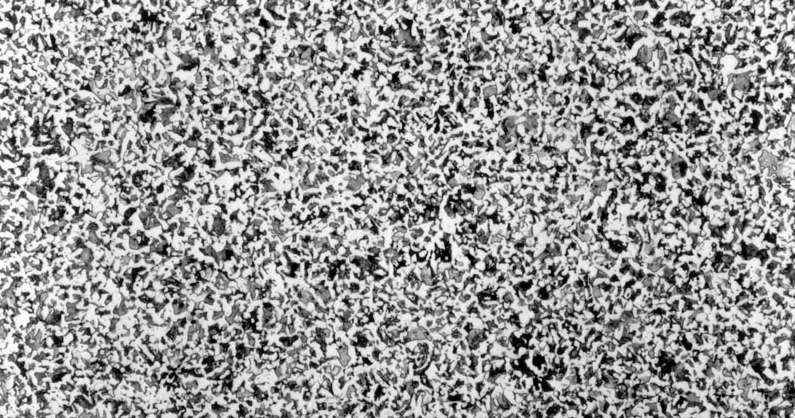}}
    {\caption{Case 2 - grayscale eutectic steel micrograph \cite{}. $795 \times 418$ pixels.}
    \label{fig:data2}}
    \end{subfigure}
\begin{subfigure}{\textwidth}
    \centering
    {\includegraphics[width=\textwidth]{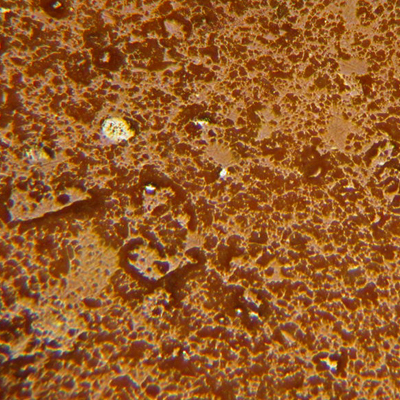}}
    {\caption{Case 3 - colour terracotta pot micrograph \cite{}. $400 \times 400$ pixels.}
    \label{fig:data3}}
    \end{subfigure}
\end{subfigure}
{\caption{Full dataset used in training for each case study.}
    \label{fig:data}}
\end{figure}

\subsection*{Seed propagation}
\label{sec:seedprop}
\begin{figure}[H]
\centering
{\includegraphics[width=\textwidth]{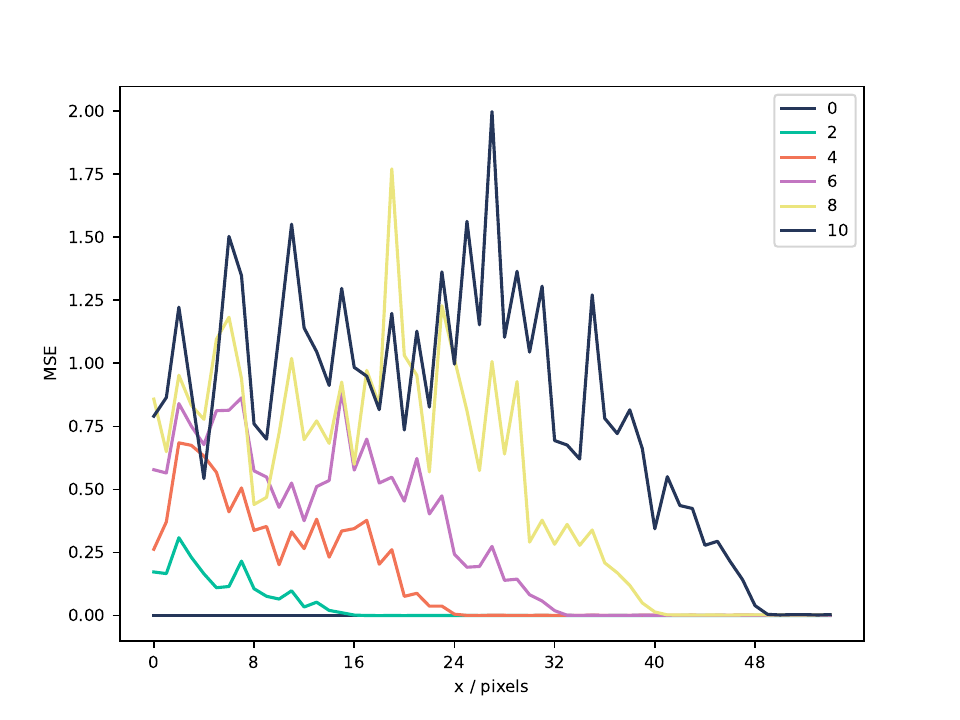}}
{\caption{A plot showing the MSE between a baseline image and a generated image with the seeds changed. The original size of the image is $112 \times 112$ pixels, generated from a $16 \times 16$ seed. The x-axis is moving away from the center of the image, and the different lines correspond to different $n\times n$ squares of seeds being changed in the center of the image.}
    \label{fig:seed}}
\end{figure}

The design of the architecture means that approximately one seed corresponds to an 8x8 region. However, this is not exactly the case and the reader is directed to \href{convolv.in}{convolv.in} to explore the way different configurations of convolutions affect the output size of the network. To visualise the effect of changing seeds on the output of the generator, a baseline sample was generated from a single random seed. Then, marching outwards, the seed is changed one step at a time, and the difference between the resulting output and the baseline is taken. This is then summed in one spatial direction, and plotted in Figure \ref{fig:seed}. The central 2x2 region affects a region approximately 32\textsuperscript{2}, each step following roughly increases the effected region by 8 pixels on each side.

The methods developed in this paper take an annulus of width 16 pixels when calculating the content loss. In order to safely change the seeds to not affect the border matching, we ensure a buffer of 8 pixels on each side, and a minimum area of 32 pixels in the center to change. Therefore, the minumum seeds size is $10 \times 10$, as this generates a $64 \times 64$ pixel image. Above this, the seeds that can be changed scale with the following formula: $n_{\Delta}=2*(n_{seed}-10)$. For example, for a $12 \times 12$ seed, a $4 \times 4$ region can be changed.

\subsection*{Volume fraction KS tests}
The p-values for the KS tests between the distribution of volume fractions for the three-phase SOFC material in case 1. To calculate these, a KS test was performed between the ground truth of each phase and the corresponding method.

\begin{table}[H]
\centering
\begin{tabular}{lllll}
\toprule
Phase & G rand. & G fixed &   Seed opt. & Seed rand. \\
\midrule
   Pore & 0.73 & 1.2e-12 & 4.6e-08 &    0.43 \\
   Metal & 0.91 & 2.3e-16 & 1.1e-45 &   0.022 \\
   Ceramic & 0.97 & 7.3e-05 & 4.9e-50 &   0.022 \\
\bottomrule
\end{tabular}
\caption{KS test p-values of the volume fraction distributions for each of the methods for unoptimised and optimised cases across each three phases.}
\label{tab:vf_ks}
\end{table}

The results clearly show G rand performs the best, achieving a less significant result than the seed rand case. Although the G fixed method performs much worse than either random method, it achieves a p-value many orders of magnitude larger than the seed opt case, indicating that the seed opt distributions are more significantly different to the ground truth.

\subsection*{Network architecture}

\begin{table}[H]
\begin{tabular}{|l|l|l|}
\hline
\textbf{Layer} & \textbf{Generator}                                 & \textbf{Discriminator}              \\ \hline
0              & TransposeConvolution(100, 512, 4, 2, 2)            & Convolution(n\textsubscript{phases}, 64, 4, 2, 1) \\ \hline
1              & TransposeConvolution(512, 256, 4, 2, 2)            & Convolution(64, 128, 4, 2, 1)       \\ \hline
2              & Convolution(256, 128, 3, 1, 1), Upsample(size*2+2) & Convolution(128, 256, 4, 2, 1)      \\ \hline
3              & Convolution(128, n\textsubscript{phases}, 3, 1, 1)               & Convolution(256, 512, 4, 2, 1)      \\ \hline
4              &                                                    & Convolution(512, 1, 4, 2, 1)        \\ \hline
\end{tabular}
\caption{Network architecture and hyperparameters for both discriminator and generator networks. The number of phases (n\textsubscript{phases}) is a parameter that varies depending on the input type. For \textit{Convolution} and \textit{TransposeConvolution} the order of the hyperparameters is \textit{(input channels, ouput channels, kernel size, stride, padding)}. For \textit{Upsample} the hyperparameter is \textit{output size}, and the method used is a bilinear interpolation.}
\label{tab:networks}
\end{table}

\subsection*{Pixel value analysis for grayscale and colour}

\begin{figure}[!h]
\centering
{\includegraphics[width=\textwidth]{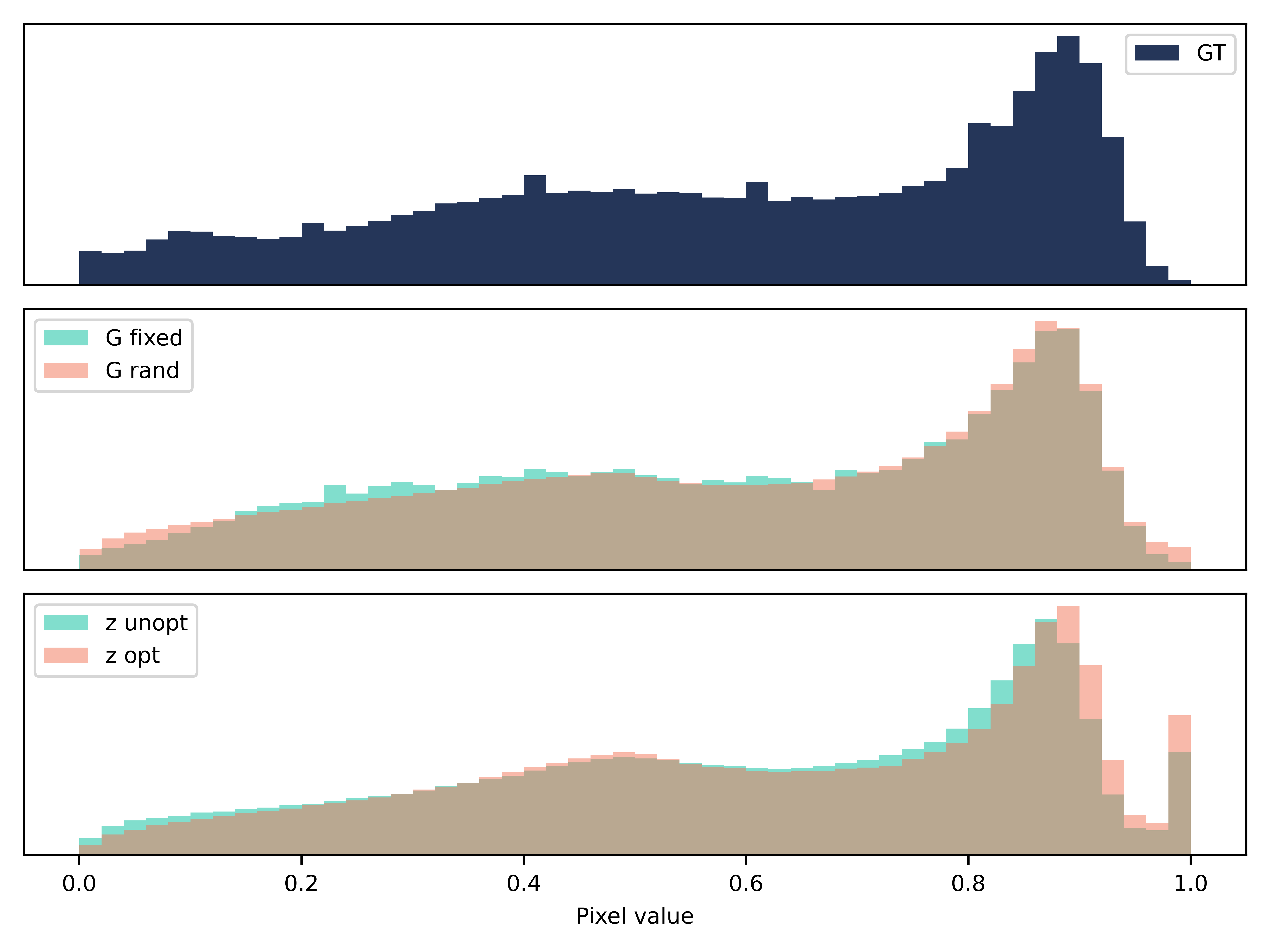}}
{\caption{Case 2: Grayscale. A histogram of pixel values for 128 samples of size $80 \times 80$ pixels. The vertical axis is the frequency of occurrence of a particular bin of pixel values.}
    \label{fig:pixel_grayscale}}
\end{figure}

\begin{figure}[!h]
\centering
{\includegraphics[width=\textwidth]{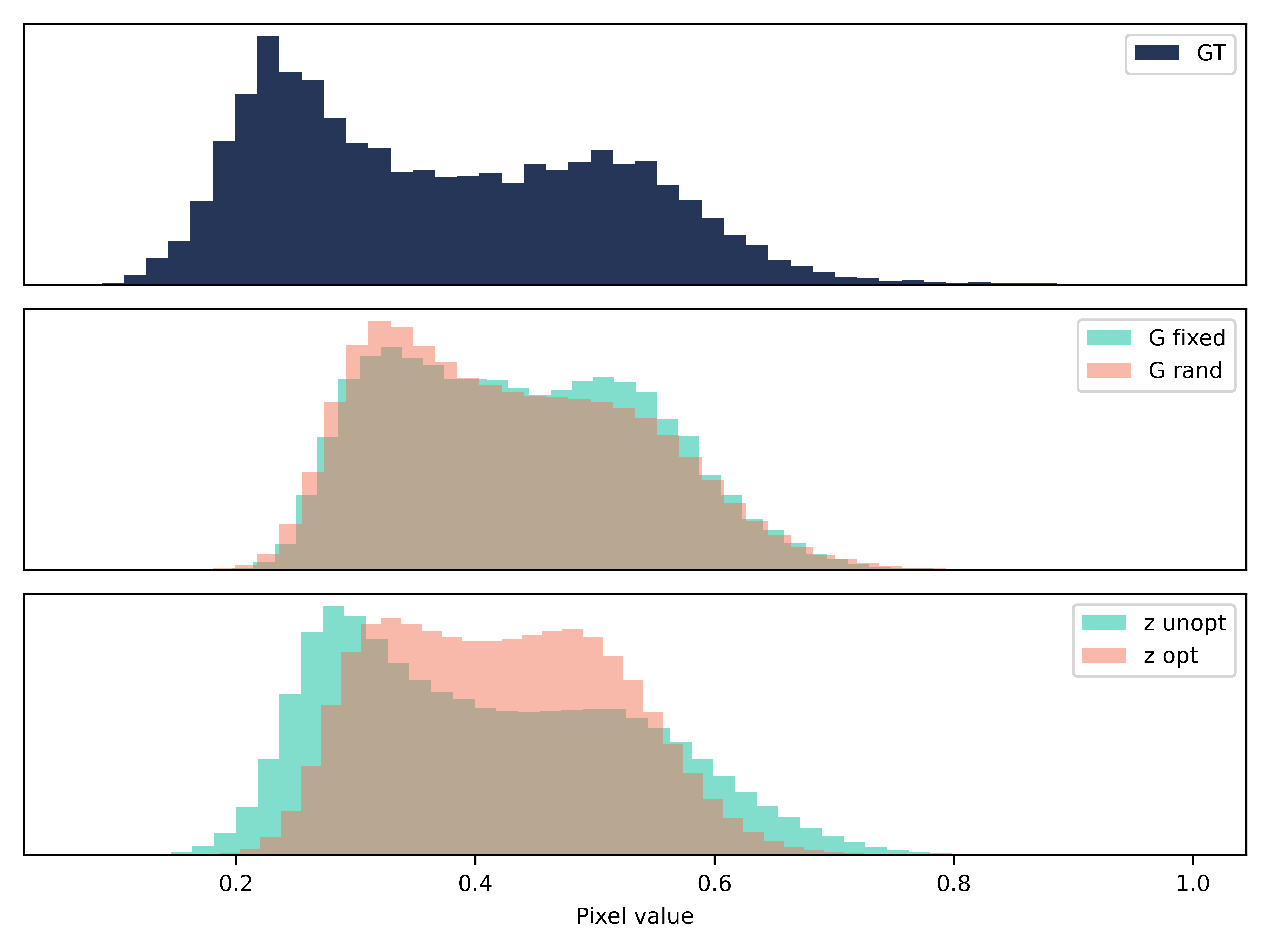}}
{\caption{Case 3: Colour. A histogram of pixel values for 128 samples of size $80 \times 80$ pixels after being transformed to grayscale. The following transformation was used: $v_{gray} = v_{colour} \cdot [0.2989, 0.5870, 0.1140]^T$. The vertical axis is the frequency of occurrence of a particular bin of pixel values.}
    \label{fig:pixel_colour}}
\end{figure}

To compare the quality of the generated images in the grayscale and colour case, we cannot compare derived microstructural metrics. Instead, we plot the distribution of pixel values and compare to the ground truth. As evident in Figure \ref{fig:pixel_grayscale}, the optimisation of the seed in Case 2 drives the generator to output more pixels with the value 1. This was reflected in Figure \ref{fig:case2}, as there appeared to be an over representation of white blobs in the microstructure.

In the colour case, we see both methods fail to replicate the distribution very well, but we can observe that the optimisation of the seed appears to flatten the central hump of the distribution, moving it further away from the ground truth.

\end{document}